\documentclass[final, custom]{anthology-ch}        

\usepackage{booktabs}
\usepackage{graphicx}

\title{Modeling turn-taking with distant viewing: investigating silence thresholds in human and AI-generated discourse}

\author[1]{Taylor Arnold}[
  orcid=0000-0003-0576-0669
]

\author[2]{Nicolas Ballier}[
  orcid=0000-0003-2179-1043
]

\author[2]{Artem Saloev}[
  orcid=0009-0006-3060-0893
]

\affiliation{1}{Data Science and Statistics, University of Richmond, Virginia, U.S.A.}
\affiliation{2}{ALTAE, Université Paris Cité, Paris, France}

\keywords[eng]{diarization, distant viewing, turn-taking, shot segmentation}

\pubyear{2026}
\pubvolume{1}
\pagestart{1}
\pageend{1}
\paperorder{1}
\conferencename{Proceedings of Conference XXX}
\conferenceeditors{Editor1 Editor2}
\doi{00000/00000}  
\customhead{Méthodes pour les sciences sociales et les humanités (2026)}

\begin{document}

\maketitle

\begin{abstract}
When one speaker stops and the next begins, a brief silence opens up. Speakers use it to manage the back-and-forth of talk and to mark whose turn it is. We measured how long that silence runs in two kinds of audiovisual material: thirty U.S. situational comedies, and fifty-one synthetic podcasts generated with Google's NotebookLM \cite{google2023notebooklm}. Gaps were compared across speaker gender, assigned from a fundamental-frequency threshold estimated in Praat \cite{boersma2024praat}, and across production setting. Two pipelines did the work. The audio pipeline, built on the pyannote suite \cite{bredin23_interspeech, plaquet23_interspeech}, recovers speaker turns and the silences between them; the image pipeline runs shot-cut detection and face-based shot-type classification over the picture track \cite{soucek2024transnet}. We report how accurately the system counts speakers in the synthetic corpus, and we lay out how gap durations split between scripted broadcast speech and synthesized dialogue. The television data tell a sharper story. Gaps that land on a shot cut run much longer than gaps inside a single shot, which means audio-only diarization quietly folds the editor's rhythm into what looks like the speakers' timing. Both registers turn out to be thin on overlap. The overlap-rich register, the spontaneous talk of people sharing a room that earlier work ties to wide speaker-to-speaker variation \cite{ng2024investigating}, sits outside this study, and reaching it is where we want to go next.
\end{abstract}

\section{Introduction}\label{sec1}
This study has two aims. The first is to find out whether automated diarization can measure the timing between speaking turns in audiovisual media. The second is to gauge how much of that timing reflects the interaction itself, as opposed to the conditions under which the source was made. The little silence between one speaker stopping and the next starting ranks among the most carefully studied quantities in conversation analysis, and for a good reason: it is easy to measure and tightly governed. People hand the floor back and forth within a few hundred milliseconds, leaving almost no gap and almost no overlap \cite{SacksSchegloffJefferson2015, HeldnerEdlund2010}. Diarization tools listen to the audio and nothing else. They recover who spoke when, with no sense of the picture or the physical setting that paces speech in broadcast media \cite{arnold2021introduction}. That makes the inter-turn gap a tempting measurement, and a slippery one. Lift it out of spontaneous talk and apply it to edited or synthesized media, and something other than the speakers may be deciding how long the silence runs.

We take the gap between turns as a compact stand-in for turn-taking and track it across two very different registers: U.S. broadcast situational comedies on one side, synthetic podcasts from Google's NotebookLM on the other \cite{google2023notebooklm}. Setting them side by side shows what a machine-generated conversation looks like next to scripted human media, and whether a generative system mimics the give-and-take timing of human talk or flattens it into something more uniform. We also bring in the picture. Running shot-cut detection and shot-type classification over the image track tells us where audio-only measurement has to be read against the way the source was put together, and it reveals that a sizable chunk of the silence we pin on speakers actually comes from the editor.

\section{Previous Research}\label{sec2}
\subsection{Turn-taking modelling}\label{subsec21}
For decades, turn-taking has been understood as a system built to keep gaps and overlaps small, with timing at its core \cite{skantze2015exploring}. The classic account describes turn allocation as locally managed and party-administered: transitions cluster around points where a turn could plausibly end, so the next speaker can come in with little or no gap \cite{SacksSchegloffJefferson2015}. Phonetic work later put numbers to all this. It teased apart real between-speaker gaps from within-speaker pauses and from overlaps, and found that the usual transition sits at a short positive offset rather than at zero \cite{HeldnerEdlund2010}. The timing holds up across wildly different languages. One large cross-linguistic study found that the exact values shift from culture to culture, yet every community in the sample steers clear of long silences and settles on quick handoffs, which hints at a shared machinery underneath \cite{StiversEnfieldBrown2009}. Previous research on NotebookLM turn-taking investigated male and female personae and potential accommodation of rhythm across personae \cite{saloev26_speechprosody}.

On the computational side, the methods run from rule- and signal-based timing models to data-driven predictors that guess where a turn will end, and recent surveys chart that range along with the wider move toward learned models and their place in spoken dialogue systems and human-robot interaction \cite{Skantze2021, castillo-lopez-etal-2025-survey}. One limitation keeps showing up. These systems work from the acoustic signal alone, treating a turn boundary as a matter of audio timing and ignoring the visual or material setting the interaction happens in. For two people talking in the same room, the shortcut does little harm, because the audio really does carry the interactional signal. For edited and produced media it harms a lot, since the timing on the soundtrack has been worked over by hands that are not the speakers'.

\subsection{Distant viewing with speech data}\label{subsec22}
Distant viewing carries the methods of distant reading and cultural analytics over to large audiovisual collections, handling the image track as data to be studied at scale next to text and audio \cite{arnold2023distant, Manovich2020}. Its roots are in computational film and media studies, where measurable features of style, things like shot length and the rhythm of the cutting, have served to pin down the signature of a director or an era \cite{Heftberger2018}. Film and television scholars settled long ago on a different view of these features: they are anything but incidental. How a scene gets cut, and how it gets shot, are deliberate elements of style that organize the way an audience experiences screen talk \cite{Bordwell2006, Butler2012}.

All of this bears directly on measuring speech. In broadcast media you have to account for where the shots change and for whether a studio audience is present, because both push speech timing around no matter what the speakers themselves decide \cite{arnold2025sitcom}. The sitcom makes the point with unusual clarity. Its performance conventions, and in the multi-camera tradition its live or canned audience, lay a pacing on the dialogue that belongs to the genre rather than to the conversation on screen \cite{Mills2009}. A diarization tool that hears only the audio can easily chalk up to interaction what really comes from production. So we pair audio diarization with the image track, running shot-cut detection and shot-type classification to test head-on whether the gaps we measure are tracking the cutting of the picture.

\section{Materials and Methods}\label{sec3}
\subsection{Data}\label{subsec31}
The human corpus gathers thirty U.S. situational comedies, more than 4,500 episodes across several decades, from the multi-camera studio shows of the 1950s up through the single-camera series that took over later on. That span gives us plenty of room to look at how aural pacing changes from one mode of production to another and from one decade to the next; the per-series and per-decade groupings below follow the production metadata logged for each series. The synthetic corpus is fifty-one podcasts made with Google's NotebookLM, collected between July and November 2025 \cite{google2023notebooklm}. Each one pairs two or more hosts. By convention one host is a higher-pitched female persona and the other a lower-pitched male persona, and that fixed two-host layout shows up plainly in the transition counts we report below. NotebookLM kept getting updated while we were collecting, so the data reflect the several model versions live during that window rather than one frozen system.

In both corpora, speaker gender was assigned by machine from a fundamental-frequency (F0) threshold, with F0 measured off the audio in Praat \cite{boersma2024praat}. This is a blunt acoustic proxy for labeling transitions and nothing more; it makes no claim about who a speaker actually is. The threshold sorts voices into a higher-F0 group and a lower-F0 group, which we call female and male to stay in step with earlier work, even as we grant the limits of a purely acoustic cut.

\subsection{Audio diarization with pyannote}\label{subsec32}
Both corpora ran through the C speaker-diarization suite \cite{bredin23_interspeech, plaquet23_interspeech}, which carves the audio into speaker turns and assigns each turn to a cluster. From that segmentation we read off turn boundaries, and for every adjacent pair of turns by different speakers we read off the inter-turn gap, the measurement this whole study turns on. For the synthetic corpus we also use the output to count how many distinct speakers a podcast contains, a quick check on whether the pipeline is splitting or merging synthesized voices too freely. The same pipeline runs over the television corpus, pulling turn boundaries and inter-turn gaps at scale across every episode.

\subsection{Image-track analysis}\label{subsec33}
For the television corpus we set the audio pipeline alongside an analysis of the image track. A deep shot-transition detector finds the cuts in the video \cite{soucek2024transnet}, and face detection then labels each shot by how many people it frames: one person (S) or several (M). Every inter-turn gap from the audio gets tagged by how close it sits to the nearest cut. Gaps within $\pm0.5$\,s of a cut count as ``near-cut''; the rest are mid-shot. When both shots on either side of a near-cut gap can be classified, we also note the shot-type transition across the cut, say S$\rightarrow$M. With that in hand we can pull apart the gaps that fall on an editing decision from the gaps buried inside a single shot, and we can ask whether a gap's length depends on the kind of cut sitting inside it.

\begin{table}[htbp]
\centering
\caption{Inter-turn gap durations in the human broadcast corpus, aggregated by gender transition, production style, and decade. All time values are reported in milliseconds (ms).}
\label{tab:silence_thresholds}
\begin{tabular}{@{}lccc@{}}
\toprule
\textbf{Category} & \textbf{Mean Gap (ms)} & \textbf{Median Gap (ms)} & \textbf{$n$} \\
\midrule
\multicolumn{4}{@{}l}{\textit{Gender Transitions}} \\
Female-Female (FF) & 1225 & 470 & 800,343 \\
Female-Male (FM)   & 1222 & 505 & 209,600 \\
Male-Female (MF)   & 1264 & 550 & 209,594 \\
Male-Male (MM)     & 1132 & 459 & 101,896 \\
\midrule
\multicolumn{4}{@{}l}{\textit{Production Style}} \\
Single-camera & 1067 & 424 & 578,519 \\
Multi-camera  & 1345 & 562 & 742,914 \\
\midrule
\multicolumn{4}{@{}l}{\textit{Decade}} \\
1950s & 1236 & 436 & 124,524 \\
1960s & 1280 & 516 & 167,695 \\
1970s & 1129 & 424 & 192,544 \\
1980s & 1370 & 631 & 123,549 \\
1990s & 1500 & 654 & 249,848 \\
2000s & 1107 & 470 & 349,191 \\
2010s & 875  & 355 & 114,082 \\
\bottomrule
\end{tabular}
\end{table}

\begin{table}[htbp]
\centering
\caption{Inter-turn gap durations in the synthetic NotebookLM corpus, aggregated by gender transition. All time values are reported in milliseconds (ms).}
\label{tab:synthetic_thresholds}
\begin{tabular}{@{}lccc@{}}
\toprule
\textbf{Category} & \textbf{Mean Gap (ms)} & \textbf{Median Gap (ms)} & \textbf{$n$} \\
\midrule
Female-Female (FF) & 634 & 550 & 3 \\
Female-Male (FM)   & 319 & 287 & 2,994 \\
Male-Female (MF)   & 337 & 310 & 2,997 \\
Male-Male (MM)     & 397 & 367 & 20 \\
\bottomrule
\end{tabular}
\end{table}

\begin{table}[htbp]
\centering
\caption{Inter-turn gap durations in the television corpus by relation to visual editing. Top: all gaps classified by proximity to a shot cut. Bottom: the near-cut gaps for which both adjacent shots could be classified, by shot-type transition across the cut (S = single-person shot, M = multi-person shot). Rows ordered by median gap. All time values in milliseconds (ms).}
\label{tab:visual_editing}
\begin{tabular}{@{}lccc@{}}
\toprule
\textbf{Category} & \textbf{Mean Gap (ms)} & \textbf{Median Gap (ms)} & \textbf{$n$} \\
\midrule
\multicolumn{4}{@{}l}{\textit{Cut proximity}} \\
Mid-shot & 1074 & 424 & 972,594 \\
Near-cut ($\pm0.5$\,s) & 1639 & 711 & 348,839 \\
\midrule
\multicolumn{4}{@{}l}{\textit{Near-cut, by shot-type transition}} \\
M $\rightarrow$ M & 1406 & 608 & 37,044 \\
S $\rightarrow$ M & 1476 & 665 & 54,136 \\
M $\rightarrow$ S & 1513 & 677 & 53,350 \\
S $\rightarrow$ S & 1503 & 699 & 139,464 \\
\bottomrule
\end{tabular}
\end{table}

\begin{table}[htbp]
\centering
\caption{Median inter-turn gap by series in the television corpus (all thirty series), ordered by median gap. All time values are reported in milliseconds (ms).}
\label{tab:series_thresholds}
\begin{tabular}{@{}lcc@{}}
\toprule
\textbf{Series} & \textbf{Mean Gap (ms)} & \textbf{Median Gap (ms)} \\
\midrule
Brooklyn Nine-Nine & 660 & 275 \\
Black-ish & 676 & 310 \\
Modern Family & 772 & 310 \\
30 Rock & 1016 & 355 \\
I Love Lucy & 1192 & 367 \\
Community & 1050 & 367 \\
All in the Family & 1182 & 378 \\
The Dick Van Dyke Show & 910 & 401 \\
The Good Place & 990 & 401 \\
Arrested Development & 756 & 401 \\
Good Times & 1125 & 401 \\
The Mary Tyler Moore Show & 1022 & 436 \\
Parks and Recreation & 918 & 447 \\
Kim's Convenience & 1026 & 459 \\
Frasier & 1208 & 470 \\
I Dream of Jeannie & 1415 & 482 \\
Fresh Off the Boat & 1159 & 482 \\
Sanford and Son & 1191 & 482 \\
The Donna Reed Show & 1290 & 527 \\
The Office (US) & 1233 & 539 \\
My Living Doll & 1272 & 585 \\
How I Met Your Mother & 1215 & 585 \\
Seinfeld & 1266 & 608 \\
Cheers & 1450 & 642 \\
Bewitched & 1541 & 665 \\
Friends & 1518 & 665 \\
Everybody Loves Raymond & 1680 & 688 \\
Living Single & 1606 & 848 \\
The Fresh Prince of Bel-Air & 1720 & 848 \\
The Big Bang Theory & 1622 & 929 \\
\bottomrule
\end{tabular}
\end{table}

\section{Results}\label{sec4}
\subsection{Diarization Accuracy}\label{subsec41}
On the synthetic corpus the pyannote pipeline got the speaker count right in 44 of 51 podcasts, or 86\%. The seven misses all went the same way: the pipeline over-segmented, inventing one or more phantom low-F0 (male-classified) speakers. It never under-counted. Since every NotebookLM podcast uses the same two-host format, these slips are the pipeline adding a speaker who is not there rather than failing to tell two real voices apart. The lopsidedness is worth a moment. The pipeline is cautious about merging voices but now and then splits a single host across two clusters, which puffs up the apparent count without ever folding the two genuine hosts into one. For the gap measurements ahead it barely registers, because the invented speakers are sparse and the cross-gender transitions that do the heavy lifting stay well populated.

\subsection{Silence Thresholds in Broadcast Television}\label{subsec42}
For a human baseline we worked through the inter-turn gaps across all thirty U.S. sitcoms (Table~\ref{tab:silence_thresholds}). Mean gaps get pulled upward by structural pauses, the scene transitions and the bits of physical comedy, so we lean on the median as the steadier read on conversational turn-taking. Both sit in the table.

Median gaps split by production style. Single-camera sitcoms come in at 424\,ms ($n=578{,}519$); multi-camera sitcoms run to 562\,ms ($n=742{,}914$), a difference of 138\,ms. One likely reason: the multi-camera format is built around a live studio audience or a laugh track, which forces performers to hold their turns and leave room for the crowd to react. The decade numbers line up with that. Median gaps peak in the 1980s (631\,ms) and 1990s (654\,ms), then drop to 470\,ms in the 2000s and 355\,ms in the 2010s, following the industry's drift toward single-camera production. The earlier decades land in the middle (1950s, 436\,ms; 1960s, 516\,ms), with a dip in the 1970s (424\,ms), so the trend over time does not move in one clean direction.

These are not effects you can pull cleanly apart, and we want to flag that up front. Production style, decade, and individual series are tangled together. A single series throws off many thousands of transitions, all in one style and one era, so the sample size that actually counts for separating the effects is nearer to the number of series, thirty, than to the number of transitions. The per-series medians in Table~\ref{tab:series_thresholds} make this vivid. They stretch from 275\,ms (\emph{Brooklyn Nine-Nine}) to 929\,ms (\emph{The Big Bang Theory}), several times the 138\,ms gap between single- and multi-camera, and series of the same production style land on opposite sides of the divide. The single-camera \emph{The Office (US)} at 539\,ms sits above the multi-camera \emph{Frasier} at 470\,ms; multi-camera \emph{Good Times} and single-camera \emph{Arrested Development} meet at 401\,ms. Variation between series, in short, drowns out the category effects in the aggregate. Splitting style from decade cleanly would mean treating series as a fixed effect, and thirty series is just too thin to give estimates anyone should trust.

Gender transitions show a smaller asymmetry that holds steady. The longest median gaps fall at male-to-female transitions (550\,ms) and the shortest at same-gender ones (male-male 459\,ms, female-female 470\,ms), with female-male sitting in between (505\,ms). Next to the production-style and editing contrasts these are small, so we take them as a mild tendency rather than a strong effect, and a tendency to handle with care, because the F0 threshold dumps roughly three-fifths of all transitions into the female-female bin ($n=800{,}343$).

\subsection{Gaps and Visual Editing}\label{subsec44}
The television corpus also lets us check whether the gaps we measure are tracking the cutting of the image rather than the interaction. Tag each gap by how near it falls to a shot cut (Table~\ref{tab:visual_editing}) and a big effect jumps out. Gaps that land on a cut, within $\pm0.5$\,s, have a median of 711\,ms; gaps that fall mid-shot sit at 424\,ms. That is a difference of 287\,ms. It dwarfs the production-style contrast at 138\,ms and the gender contrasts at $\le 91$\,ms, and it runs about as wide as the whole decade range (355--654\,ms). Put plainly: a good slice of what audio-only diarization clocks as a long inter-turn ``silence'' is the editor dropping a cut into the gap, not the speakers pausing. Formats and eras also differ in how often they cut, so part of the camera and decade differences above may itself be a cut-density effect. The near-cut analysis pins that editing component down on its own.

Break the near-cut gaps down by the shot types flanking the cut and the picture gets sharper. The shortest near-cut gaps show up when both shots frame several people (M$\rightarrow$M, 608\,ms); the longest come when the cut stitches together two single-person shots (S$\rightarrow$S, 699\,ms), with the mixed cases in between. This is just the grammar of dialogue editing at work. Shot/reverse-shot cutting between two people leaves a little air around the cut, while a held group framing lets the exchange move faster. The ordering fits the larger argument: editing rhythm, not interactional timing by itself, sets the silences a diarizer hauls out of the soundtrack.

\subsection{Silence Thresholds in Synthetic Speech}\label{subsec43}
Table~\ref{tab:synthetic_thresholds} gives the same gender measurements for the synthetic corpus. Two cross-gender transitions carry the data (FM, $n=2{,}994$; MF, $n=2{,}997$), exactly what NotebookLM's two-host format predicts. Same-gender transitions barely occur (FF, $n=3$; MM, $n=20$), so we leave them uninterpreted.

Synthetic gaps come in well under the broadcast medians. The cross-gender medians are 287\,ms for female-to-male and 310\,ms for male-to-female, against 505\,ms and 550\,ms in the television corpus. The \emph{direction} of the gender asymmetry survives, with male-to-female running a touch longer than female-to-male in both registers, but its size, along with the overall length of the gaps, gets squeezed down in the synthetic data. None of the production-driven inflation from broadcast speech turns up here. There is no studio audience or laugh track to play to, and no editor laying a cut into the pause to stretch it out. What is left looks close to a floor on transition timing: a near-mechanical handoff between two voices, stripped of the variability that editing and performance pour into the broadcast side.

\section{Discussion}\label{sec5}
\subsection{The Materiality of the Gap}\label{subsec51}
The gap between the two corpora reads best through the material conditions of each, and the television data show it without much help. Gaps that fall on a shot cut run a median of 287\,ms longer than gaps inside a shot, so a large share of the ``silence'' pulled from broadcast audio is a product of editing, not of interaction. Set that beside the gender and production-style contrasts, every one of them smaller, and the implication is that the picture track, and not the speakers alone, sets much of the rhythm a signal-only diarizer ends up measuring. That squares with an old claim in film and television scholarship: editing and camera setup are building blocks of screen style, not neutral pipes for the action they record \cite{Bordwell2006, Butler2012}. 

Synthetic gaps are a different. They are short and machine-regular, with none of this weighing on them, because nothing in the generation process supplies the weight. The synthetic medians, roughly 287--310\,ms for the well-populated transitions, fall below every aggregate broadcast category; the fastest of those are the single-camera format at 424\,ms and the 2010s at 355\,ms. NotebookLM, then, treats turn-taking as a near-instant handoff rather than the negotiated, context-sensitive timing of human-made media. The synthetic values even drop to or below the short positive offsets that mark unedited human conversation \cite{HeldnerEdlund2010}, so the system is not really reproducing human timing. It is squashing it toward a minimum.

\subsection{Overlap and the Limits of These Data}\label{subsec52}
The biggest limitation here is overlap. Real spoken interaction is full of it, with people coming in fast and sometimes talking at once, and telling gaps, pauses, and overlaps apart has been a methodological worry in the phonetics of conversation for a while now \cite{HeldnerEdlund2010}. Work on spontaneous-speech corpora finds a matching spread of variation between speakers, some of it driven by interactional stance \cite{ng2024investigating}, and the cross-linguistic record says that quick, low-silence transitions are the human default rather than a quirk of one setting \cite{StiversEnfieldBrown2009}. Neither corpus here reaches that register. Scripted, edited television keeps overlap out by design, and the synthetic podcasts hold almost none, because NotebookLM does not produce overlapping turns at all. So both datasets are overlap-poor. Our gap measurements describe two fairly orderly registers and have nothing to say about the overlap-rich churn of natural conversation. Reaching for a spontaneous, in-person spoken corpus was in the plan from the start, and it is still the next step that matters most.

\section{Conclusion}\label{sec6}
We measured inter-turn silence thresholds across two audiovisual registers, U.S. broadcast sitcoms and NotebookLM synthetic podcasts. In scripted television, gap durations turn out to depend on production format and era, and most plainly of all on visual editing. The synthetic dialogue does something else, putting out much shorter and more uniform gaps with little of that structure and basically no overlap. The comparison can only be partial. Both registers are overlap-poor, and in the television data the production and temporal effects are confounded at the level of the series. Taken in modest terms, the results back a plain methodological lesson for distant viewing of audiovisual media. Gap timing recovered by signal-only diarization reflects the material and production conditions of the source, the placement of shot cuts above all, every bit as much as it reflects the interactional choices of the speakers, and treating it as turn-taking ``strategy'' means accounting for those conditions first. The obvious way forward is to bring spontaneous, overlap-rich speech into the comparison, which would let the orderly registers studied here be set against the negotiated timing that turn-taking research has charted in unscripted human talk.

\printbibliography

\end{document}